\pdfoutput=1
\documentclass[11pt]{article}

\usepackage[final]{acl}
\usepackage{times}
\usepackage{latexsym}

\usepackage[T1]{fontenc}
\usepackage[utf8]{inputenc}
\usepackage{amsmath}
\usepackage{microtype}

\usepackage{inconsolata}

\usepackage{graphicx}
\usepackage{booktabs}
\usepackage{multirow}
\usepackage{enumitem}

\title{Small Encoders Can Rival Large Decoders in Detecting Groundedness}

\author{
    Istabrak Abbes$^{1, 2, 3}$ \quad
    Gabriele Prato$^{1, 2, 3}$ \quad
    Quentin Fournier$^{2}$ \\
    \textbf{Fernando Rodriguez$^{5}$} \quad
    \textbf{Alaa Boukhary$^{5}$} \quad
    \textbf{Adam Elwood$^{5}$} \quad
    \textbf{Sarath Chandar}$^{1,2,4,6}$\\
    $^1$Chandar Research Lab \quad 
    $^2$Mila – Quebec AI Institute \quad
    $^3$Université de Montréal \\
    $^4$Polytechnique Montréal  \quad
    $^5$Aily Labs \quad
    $^6$Canada CIFAR AI Chair \\
    \texttt{\{firstname.lastname\}@mila.quebec} \quad
    \texttt{\{firstname.lastname\}@ailylabs.com}
}

\begin{document}

\maketitle
\begin{abstract}
	Augmenting large language models (LLMs) with external context significantly improves their performance in natural language processing (NLP) tasks. However, LLMs struggle to answer queries reliably when the provided context lacks information, often resorting to ungrounded speculation or internal knowledge. Groundedness -- generating responses strictly supported by the context -- is essential for ensuring factual consistency and trustworthiness. This study focuses on detecting whether a given query is grounded in a document provided in context before the costly answer generation by LLMs. Such a detection mechanism can significantly reduce both inference time and resource consumption. We show that lightweight, task-specific encoder models such as RoBERTa and NomicBERT, fine-tuned on curated datasets, can achieve accuracy comparable to state-of-the-art LLMs, such as Llama3 8B and GPT4o, in groundedness detection while reducing inference latency by orders of magnitude.
    The code is available at : https://github.com/chandar-lab/Hallucinate-less
\end{abstract}

\section{Introduction}
\label{sec:Introduction}
Large language models have demonstrated remarkable capabilities in various tasks, from text generation~\citep{zhao_survey_2024} to question answering~\citep{zhao_survey_2024}. However, their tendency to hallucinate when the query lacks support from the provided context or when the model faces noisy retrieval~\citep{yoran_making_2024, wu_how_2024} raises concerns regarding their reliability~\citep{ji_survey_2023}. Ideally, LLMs should only answer if the provided context contains enough information to answer the question when combined with the model’s parametric knowledge. Otherwise, the model should abstain from answering or ask for more information. Some works have investigated this issue by evaluating RAG-LLMs in the presence of irrelevant information in the context~\citep{wang_rear_2024,cuconasu_power_2024,xie_adaptive_2024}.

However, these works lack a consistent definition of ``relevant information''. It could refer to anything from content that directly answers the question to material that is only loosely associated with the topic. This study defines relevance as information that directly contributes to answering the question and frames the problem as a classification task that distinguishes between relevant and irrelevant information. Encoders, such as BERT~\citep{devlin_bert_2019}, RoBERTa~\citep{liu_roberta_2019}, and NomicBERT~\citep{nussbaum_nomic_2025}, have demonstrated strong capabilities in capturing contextual relationships and producing high-quality embeddings for downstream tasks, which can be useful for factual verification. On the other hand, decoders such as Llama have demonstrated strong capabilities in open-ended text generation. However, they often require significantly more computational resources as shown in Table \ref{tab:acc}. We evaluate the ability of these models to measure groundedness, defined as whether the document supports the query, irrespective of the model’s prior knowledge. This evaluation is conducted on encoder-based and decoder-based models using a variety of question-answering (QA)~\citep{rajpurkar_know_2018,trischler_newsqa_2017} and information retrieval datasets~\citep{thakur_beir_2021}. 

Our analysis reveals two key findings:
\begin{itemize}[itemsep=2pt,parsep=2pt,partopsep=2pt,topsep=2pt]
	\item Fine-tuned encoders can achieve comparable accuracy to the best LLMs in groundedness detection while reducing inference costs by up to 1,000x.
	\item The zero-shot performance of LLMs in groundedness detection greatly depends on the prompt, especially for smaller models.
\end{itemize}

Our findings suggest that practitioners can utilize encoders for groundedness detection in their workflows, achieving similar efficacy to LLMs while significantly decreasing computing costs.

\section{Related Work}
\label{sec:Related Work}

A recurring issue in generative models is hallucinations, where the model produces information that appears plausible but is not grounded in reality. \citet{ji_survey_2023} provide a comprehensive taxonomy of hallucinations, categorizing them into intrinsic when arising from model biases, and extrinsic when resulting from data limitations. The former occurs when the generated text includes factual inaccuracies or entities not present in the input, while the latter occurs when the model generates content not supported by the available data or context.

Groundedness in NLP refers to a model's ability to generate outputs consistent with factual knowledge and the given input context. Retrieval-augmented generation (RAG)~\citep{lewis_retrieval-augmented_2021} improves groundedness by incorporating external sources during inference. However, given that RAG leads to very long contexts, the  ``lost in the middle'' problem~\citep{hsieh_found_2024,liu_lost_2023} appears, where models struggle to use information positioned in the middle of long input sequences. Some works have tried to solve the problem by focusing on improving the generation and retrieval quality, often by fine-tuning one or more components~\citep{asai_self-rag_2023,zhang_end--end_2024}. However, those works consider relevant documents only for their analysis. As RAG systems surface top-ranked documents, they can still include irrelevant distractions~\citep{cuconasu_power_2024,asai_reliable_2024,wang_learning_2023}. 
% \citet{wu_instructing_2024}  add specific instructions into prompts, verifying the irrelevant content within problem descriptions. However, a small change in the prompt can have an effect on the performance~\citep{salinas_butterfly_2024}.

Several methods employ a model to predict relevance scores within a larger pipeline \citep{wang-etal-2024-rear,zhou2024metacognitiveretrievalaugmentedlargelanguage,jiang2025retrievesummarizeplanadvancing}
but none of them tried encoders for relevance detection.

\begin{figure}[!t]
	\includegraphics[width=\columnwidth]{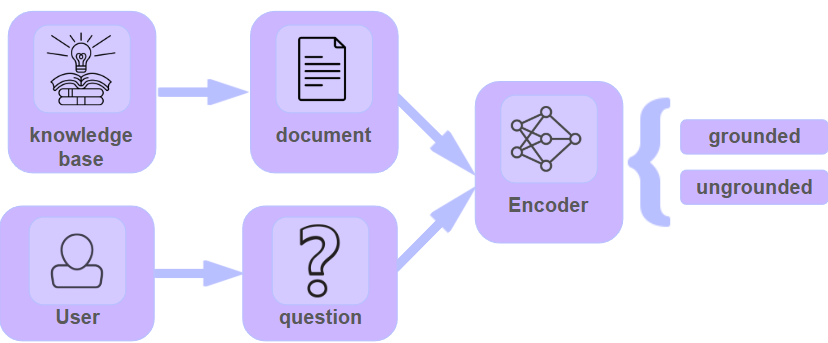}
	\caption{Detecting groundedness before passing the information to the LLM to avoid hallucinations and reduce computational costs.}
	\label{fig:paper}
\end{figure}

\section{Experimental Setup}
\label{sec:Experimental Setup}

Our experiments aim to evaluate the ability of encoder-based models to determine whether a given query is grounded in the provided context before engaging in the computationally expensive process of answer generation by large language models.

We work with a dataset \( D \), consisting of instances \( q = (Q, C) \), where \( Q \) represents the query, \( C \) represents the provided context, which may or may not contain sufficient information to answer \( Q \). The objective is to train a model that can classify each \( (Q, C) \) pair into \textit{relevant} or \textit{irrelevant}, based on whether the context provides enough support for answering the query.

During training and inference, we don't answer the question, as our goal is to detect irrelevant context before answering to avoid unnecessary computation.
In other words, our approach complements existing retrieval mechanisms rather than substituting them. Specifically, our encoder-based groundedness detection step assumes that contexts have already been retrieved and serves as a subsequent filtering mechanism. This step aims to prevent unnecessary computational overhead by ensuring that expensive inference by large language models (LLMs) is only invoked when contexts are sufficiently grounded.

\subsection{Datasets}

To systematically evaluate the groundedness capabilities of encoder-based and decoder-based models, we conducted experiments on diverse datasets covering two key NLP domains: question answering (QA) and information retrieval (IR). 

\paragraph{Question Answering} For the QA task, we utilized the SQuAD v2.0 ~\citep{rajpurkar_know_2018} and NewsQA datasets ~\citep{trischler_newsqa_2017}, both of which present a challenging setup that includes both answerable and unanswerable questions. These datasets require models to effectively discern whether the provided context contains sufficient information to answer the given query. Specifically, SQuAD v2.0 introduces unanswerable questions that necessitate precise comprehension of the passage to avoid generating unsupported answers. NewsQA provides complex questions derived from news articles, requiring a deeper understanding of the context to determine whether an answer can be inferred.

\paragraph{Information Retrieval} To extend the evaluation of groundedness beyond the QA paradigm, we leverage two subsets from the BEIR benchmark ~\citep{thakur_beir_2021}: TREC-COVID~\citep{thakur_beir_2021} and Touché~\citep{thakur_beir_2021}. The TREC-COVID dataset focuses on biomedical literature related to COVID-19, offering a realistic scenario where models must retrieve relevant scientific documents based on complex queries. It challenges the model's ability to assess the sufficiency and relevancy of retrieved information in a high-stakes domain. In contrast, the Touché dataset addresses argument retrieval, requiring models to find documents that provide argumentative support or refutation for controversial topics.

By considering these datasets, our study provides a comprehensive evaluation of groundedness across both QA and IR tasks, enabling a thorough analysis of the strengths and limitations of encoder- and decoder-based models in different scenarios. 

\subsection{Methodology}

In our experiments, we assessed the groundedness capabilities of both encoders such as BERT, RoBERTa, and Nomic-BERT, and decoder-based models, such as GPT-4 and Llama-3, across different configurations, including fine-tuned and zero-shot settings. 

\paragraph{Encoders} Encoders were fine-tuned using supervised learning to classify context-question pairs as either grounded or ungrounded. The input format for these models involved concatenating the context and question with a designated separator token, such as \texttt{[SEP]} for BERT-based architectures. Additionally, the \texttt{[CLS]} token is used to classify groundedness. We performed a hyperparameter grid search on the learning rate, weight decay, and batch size (details in Appendix~\ref{sec:hyperparameter}). 

\paragraph{Decoders} Decoders were assessed in both zero-shot and fine-tuned settings. In the zero-shot setting, we leveraged carefully designed prompts to elicit binary groundedness judgments. To minimize ambiguity and ensure consistency, we employed structured prompt templates, such as ``\texttt{Given the provided context, is the question answerable using only the information from the context? Respond with yes or no. Provide no explanation.}''

To comprehensively assess decoder model performance, we tested 20 carefully optimized prompts for QA datasets (SQuAD v2.0, NewsQA) and 20 tailored prompts for IR datasets (TREC-COVID, Touché) to enhance zero-shot accuracy and relevance evaluation (Appendix~\ref{sec:prompts}).

Additionally, we fine-tuned Llama-3 1B, 3B, and 8B models on the groundedness detection task. In this supervised fine-tuned configuration, the model was trained using labeled query-context pairs to improve its performance in identifying groundedness. 
We evaluated all models using accuracy to measure their classification performance. Results are reported as the mean and standard deviation across five random seeds.

\section{Results and Discussion}
\label{sec:Results and Discussion}
\begin{table*}[htb]
	\centering
	\resizebox{\linewidth}{!}{%
		\begin{tabular}{llcc|cc|cc|cc|cc}
			\toprule
			& & \multicolumn{2}{c|}{SQuADv2} & \multicolumn{2}{c|}{NewsQA} & \multicolumn{2}{c|}{TREC-COVID} & \multicolumn{2}{c|}{Touché} & \multicolumn{2}{c}{FLOPs} \\
			& & 0-Shot & FT & 0-Shot & FT & 0-Shot & FT & 0-Shot & FT & FT & Inference \\
			\midrule
			\multirow{9}{*}{\textbf{Encoder}} 
			& BERT-base            & -    & 64.1 {\footnotesize $\pm$ 0.4} & -    & 85.7 {\footnotesize $\pm$ 0.4} & -    & 72.9 {\footnotesize $\pm$ 0.1} & -    & 77.6 {\footnotesize $\pm$ 0.3} & $9.3 \times 10^{11}$ & $3.1 \times 10^{11}$ \\
			& BERT-large           & -    & 68.8 {\footnotesize $\pm$ 0.8} & -    & 86.1 {\footnotesize $\pm$ 0.4} & -    & 62.8 {\footnotesize $\pm$ 1.3} & -    & 79.6 {\footnotesize $\pm$ 1.4} & $2.5 \times 10^{12}$ & $8.5 \times 10^{11}$ \\
			& RoBERTa-base         & -    & 75.8 {\footnotesize $\pm$ 0.3} & -    & 85.8 {\footnotesize $\pm$ 0.6} & -    & 73.6 {\footnotesize $\pm$ 0.3} & -    & 79.3 {\footnotesize $\pm$ 0.3} & $1.1 \times 10^{12}$ & $3.7 \times 10^{11}$ \\
			& RoBERTa-large        & -    & \textbf{90.2 {\footnotesize $\pm$ 0.4}} & -    & 88.5 {\footnotesize $\pm$ 0.3} & -    & 75.7 {\footnotesize $\pm$ 0.7} & -    & 79.2 {\footnotesize $\pm$ 0.2} & $3.3 \times 10^{13}$ & $1.1 \times 10^{12}$ \\
			& Nomic-BERT           & -    & 79.8 {\footnotesize $\pm$ 0.5} & -    & 88.1 {\footnotesize $\pm$ 0.4} & -    & 74.2 {\footnotesize $\pm$ 0.6} & -    & \textbf{82.4 {\footnotesize $\pm$ 0.6}} & $9.3 \times 10^{11}$ & $3.1 \times 10^{11}$ \\
			& SimCSE-RoBERTa-large & -    & 88.3 {\footnotesize $\pm$ 0.2} & -    & 85.9 {\footnotesize $\pm$ 1.3} & -    & 72.0 {\footnotesize $\pm$ 2.0} & -    & 80.8 {\footnotesize $\pm$ 1.6} & $3.3 \times 10^{13}$ & $1.1 \times 10^{12}$ \\
			& NeoBERT              & -    & 79.0 {\footnotesize $\pm$ 0.3} & -    & 86.2 {\footnotesize $\pm$ 0.2} & -    & 73.8 {\footnotesize $\pm$ 0.2} & -    & 79.6 {\footnotesize $\pm$ 0.3} & $1.8 \times 10^{12}$ & $5.9 \times 10^{11}$ \\
			& ModernBERT-base      & -    & 86.4 {\footnotesize $\pm$ 0.2} & -    & \textbf{89.2 {\footnotesize $\pm$ 0.3}} & -    & \textbf{75.9 {\footnotesize $\pm$ 0.1}} & -    & 81.3 {\footnotesize $\pm$ 0.3} & $1.5 \times 10^{12}$ & $5.1 \times 10^{11}$ \\
			& LLM2Vec              & -    & 62.1 {\footnotesize $\pm$ 0.4} & -    & 85.7 {\footnotesize $\pm$ 0.4} & -    & 73.9 {\footnotesize $\pm$ 0.2} & -    & 78.7 {\footnotesize $\pm$ 0.4} & -                    & - \\
			\midrule
			\multirow{3}{*}{\textbf{Open LLM}} 
			& LLaMA-3.2-1B-Instruct & 55.2 & 56.8 {\footnotesize $\pm$ 1.4} & 55.3 & 84.0 {\footnotesize $\pm$ 1.3} & 37.5 & 50.2 {\footnotesize $\pm$ 0.4} & 40.8 & 48.5 {\footnotesize $\pm$ 0.8} & $1.1 \times 10^{16}$ & $2.9 \times 10^{12}$ \\
			& LLaMA-3.2-3B-Instruct & 75.9 & 82.2 {\footnotesize $\pm$ 0.8} & 78.9 & 86.4 {\footnotesize $\pm$ 0.6} & 71.4 & 74.2 {\footnotesize $\pm$ 1.1} & 76.3 & 82.4 {\footnotesize $\pm$ 1.4} & $2.1 \times 10^{16}$ & $7.0 \times 10^{12}$ \\
			& LLaMA-3.1-8B-Instruct & 81.9 & \textbf{91.1 {\footnotesize $\pm$ 0.8}} & 79.4 & \textbf{92.3 {\footnotesize $\pm$ 0.3}} & 73.8 & \textbf{75.5 {\footnotesize $\pm$ 0.6}} & 75.4 & \textbf{86.1 {\footnotesize $\pm$ 0.6}} & $6.1 \times 10^{16}$ & $1.6 \times 10^{13}$ \\
			\midrule
			\multirow{2}{*}{\textbf{Closed LLM}} 
			& Claude 3.5 V2         & \textbf{92.5} & - & \textbf{96.7} & - & \textbf{79.2} & - & \textbf{85.4} & - & - & - \\
			& GPT-4o                & \textbf{95.5} & - & \textbf{98.1} & - & 76.1 & - & 84.9 & - & - & - \\
			\bottomrule
		\end{tabular}
	}
	\caption{Accuracy of encoders and LLMs in groundedness detection for zero-shot and fine-tuned (FT) settings. Fine-tuned encoders perform closely to the best zero-shot LLMs, except on the challenging Touché dataset.}
	\label{tab:acc}
\end{table*}

% \caption{Accuracy of encoders and LLMs in groundedness detection for zero-shot and fine-tuned settings. Fine-tuned encoders perform closely to the best zero-shot LLMs, except on the challenging Touché dataset.}
% \label{tab:acc}
% \end{table*}

% We begin by investigating how robust decoders are to prompting for groundedness detection tasks ( Appendix~\ref{sec:appendix}.

% \begin{figure}[htb]
%   \includegraphics[width=\columnwidth]{squad.png}
%   \caption{Zero-shot groundedness performance of various Llama models for various prompt templates on SquadV2.0 dataset.}
%   \label{fig:prompt_squad}
% \end{figure}

\begin{figure}[htb]
	\includegraphics[width=\columnwidth]{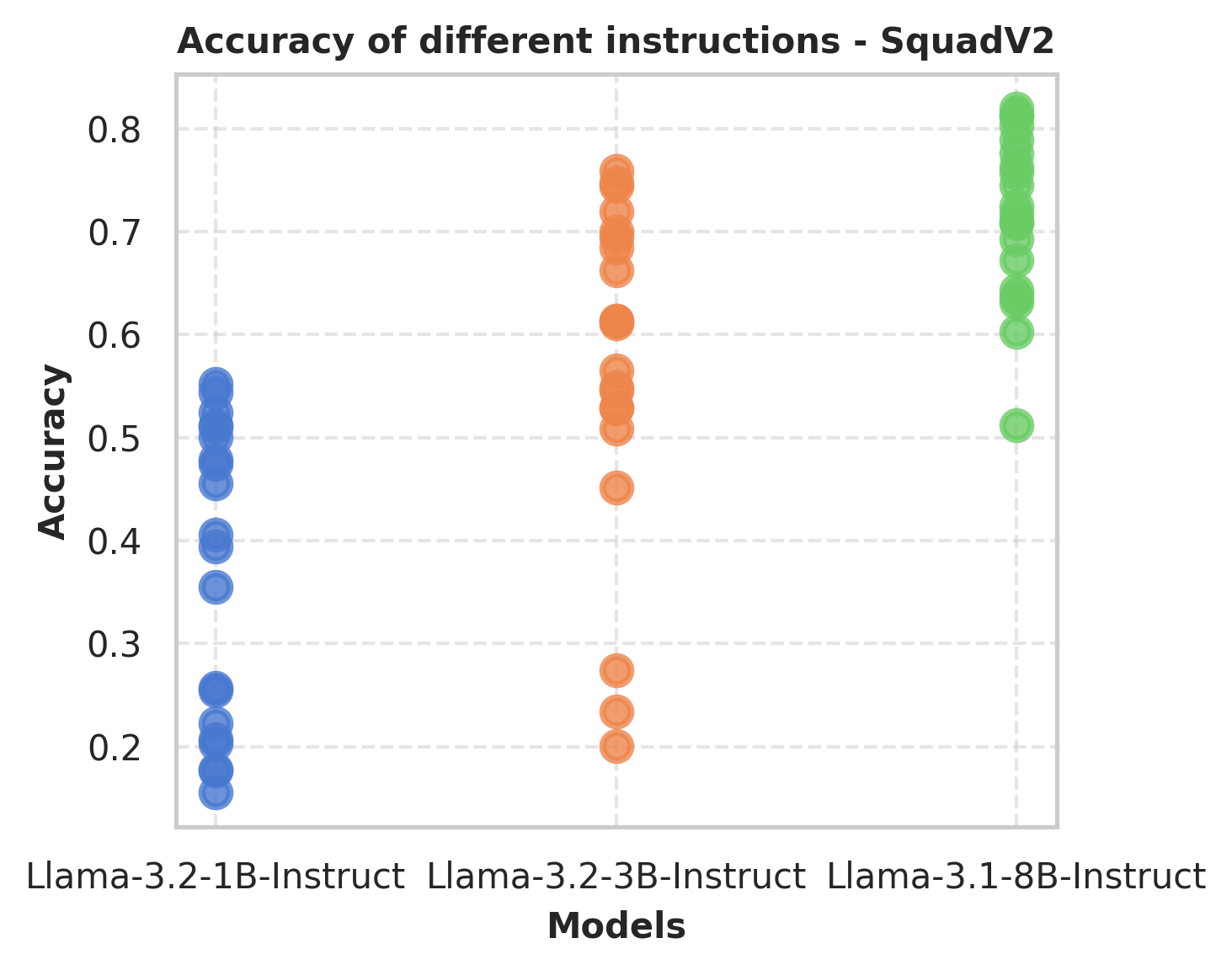}
	\caption{Zero-shot groundedness performance of various Llama models for various prompt templates on SquadV2.0 dataset.}
	\label{fig:prompt_squad}
\end{figure}

\begin{figure}[htb]
	\includegraphics[width=\columnwidth]{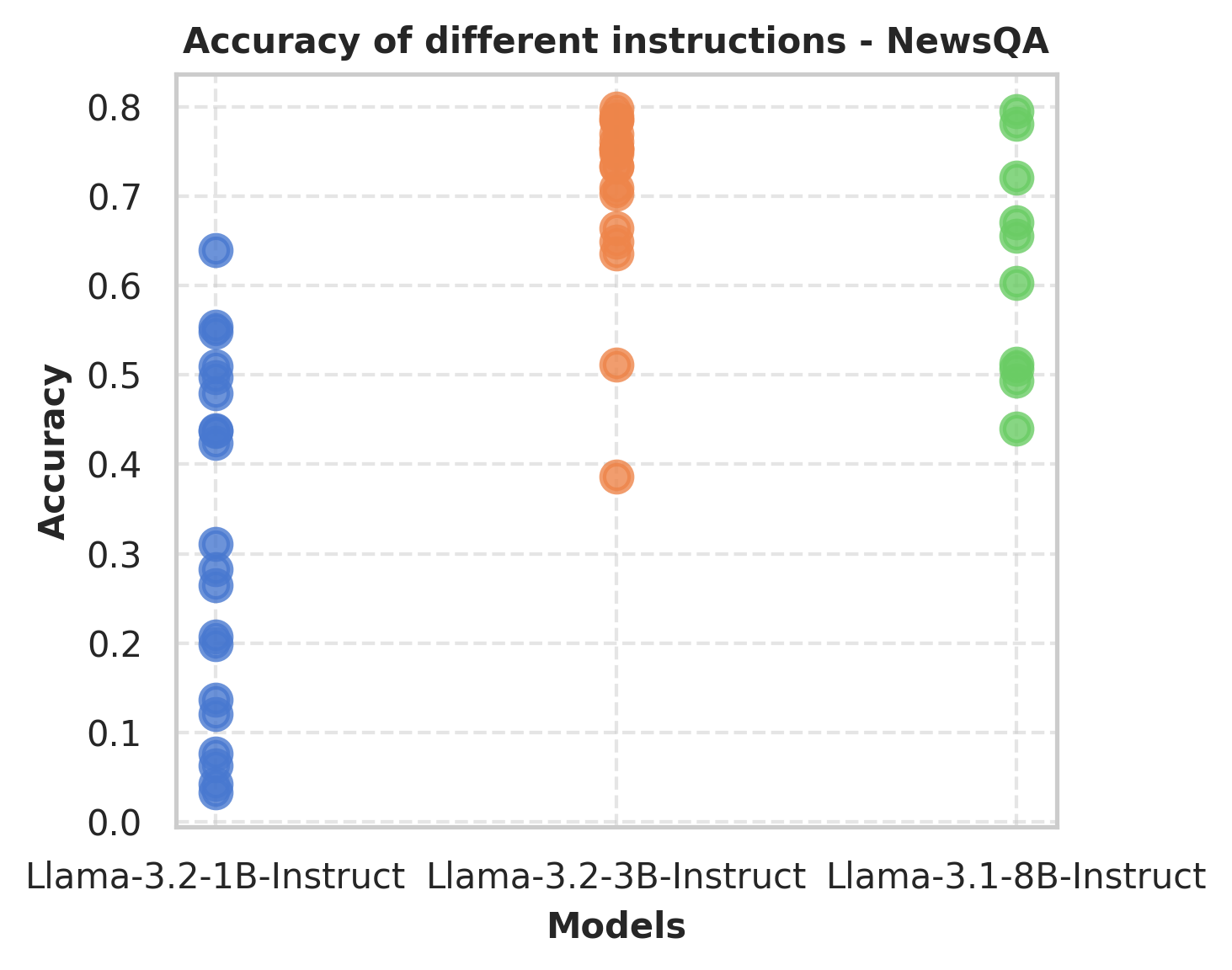}
	\caption{Zero-shot groundedness performance of various Llama models for various prompt templates on NewsQA dataset.}
	\label{fig:prompt_news}
\end{figure}

\paragraph{Interplay between Model Scale and Prompt} 
% Figures~\ref{sec:appendix} show the accuracy of different prompt instructions for models on the SQuADV2 datasets, with each point representing a distinct prompt. Models are grouped by size, with colored points indicating instructions for Llama-3.1-8B-Instruct, Llama-3.2-3B-Instruct, and Llama-3.2-1B-Instruct.
The 8B model consistently outperforms the smaller ones, suggesting that larger models, combined with specific training instructions, improve accuracy in groundedness detection. Specifically, Llama-3.1-8B-instruct achieves the highest zero-shot performance among open LLMs, with 81.9\% on SquadV2, 79.4\% on NewsQA, 73.8\% on TREC-COVID, and 74.5\% on Touché. In contrast, Llama-3.2-1B-instruct, the smallest model, performs significantly lower, with 55.2\% on SquadV2, 55.3\% on NewsQA, 37.5\% on TREC-COVID, and 40.8\% on Touché (Figures~\ref{fig:prompt_squad} and ~\ref{fig:prompt_news}).
% The 8B model consistently outperforms the smaller, suggesting that larger models, combined with specific training instructions, improve accuracy in groundedness detection. Furthermore, Llama-3.1-8B exhibits more stable performance across prompts, while smaller models, like Llama-3.2-1B, are more sensitive to prompt variations ~\citep{salinas_butterfly_2024}. 
This highlights the interplay between model scale and prompt dependency in optimizing performance, where larger models can understand the intent more reliably across phrasing  ~\citep{salinas_butterfly_2024}.

\paragraph{Impact of Model Scale and Fine-Tuning} 
% In table~\ref{tab:acc}, we summarize the performance of the models in terms of their accuracy for different configurations.

Results for different models presented in Table~\ref{tab:acc} reveal that fine-tuned encoder models, particularly RoBERTa, excel at groundedness detection, with RoBERTa-large achieving 90.2\% on SQuAD v2.0 and 88.5\% on NewsQA. While Llama3 8B achieves strong zero-shot performance, fine-tuned encoder-based models like RoBERTa still outperform it. In fact, RoBERTa large is better than the zero-shot Llama models and in between the fine-tune 3B and 8B models. This highlights that while large closed-source models like GPT-4 and Claude 3.5 V2 outperform every other models in zero-shot tasks as GPT-4o achieve 95.5\% on SQuAD v2.0 and 98.1\% on NewsQA, their results are within approximately 10\% of the best fine-tuned encoder models, which offer a more compute-efficient alternative.

Fine-tuned models, whether encoder-based or decoder-based, consistently outperform zero-shot counterparts, demonstrating the importance of task-specific training. On average, fine-tuning improves performance by 10-30 percentage points, depending on the dataset and model size. For instance, Llama-3.1-8B-instruct improves from 81.9\% (zero-shot) to 91.1\% (fine-tuned) on SQuAD v2.0 and from 79.4\% to 92.3\% on NewsQA, a boost of approximately 20 percentage points. Similarly, Llama-3.2-3B-instruct improves by about 15-20 percentage points across datasets. The gap is consistent across model types, emphasizing that pretraining alone is insufficient for optimal performance. Encoder models like RoBERTa-large (90.2\% on SQuAD v2.0, 88.8\% on NewsQA) provide comparable fine-tuned performance at a lower computational cost, making them an efficient alternative for groundedness detection in large-scale queries.

% Fine-tuned models, whether encoder-based or decoder-based, consistently outperform zero-shot counterparts, demonstrating the importance of task-specific training. Encoder models, like RoBERTa, provide similar accuracy with lower computational cost, making them ideal for groundedness detection in large-scale queries.

\paragraph{Impact of Model Size} 
Larger models generally achieve better performance, but the extent of improvement varies between encoder-based and decoder-based architectures. Among encoder models, RoBERTa-Large (355M parameters) outperforms RoBERTa-Base (125M parameters) by an average of 12-15 percentage points across datasets, highlighting the benefits of increased model capacity in fine-tuned settings. Similarly, for decoder-based models, Llama 8B instruct significantly surpasses Llama 1B instruct in zero-shot settings, with an average accuracy of 74.8\% compared to 42.1\%, an improvement of over 30 percentage points.

These results demonstrate that scaling up improves performance consistently, but smaller, fine-tuned encoder models like RoBERTa-Large can still compete with much larger decoder models.This tradeoff between model size and efficiency aligns with findings from ~\citep{zhang_sentiment_2023, zimmerman_two-tiered_2024}, which emphasize the advantages of compact, well-trained encoders in classification tasks. We believe lightweight encoders perform well at groundedness detection because the task itself, determining if a passage supports answering a query, aligns naturally with semantic matching, which is a strength of encoders trained with contrastive or classification objectives. In contrast, LLMs are optimized for open-ended generation and often rely on prompt sensitivity or parametric knowledge, which may not be necessary for binary relevance classification. Thus, in tasks with localized grounding, encoders are a better fit both computationally and inductively.

% RoBERTa-Large, a 355M parameter encoder-based model, outperforms the 3B parameter decoder-only Llama across all datasets tested, achieving an average accuracy of 83.4\% compared to 70.85\% for zero-shot Llama-3B.  These findings are consistent with~\citep{zhang_sentiment_2023, zimmerman_two-tiered_2024} which highlight the shortcomings of decoder-based LLMs for classification tasks that smaller encoder models excel in.

\paragraph{Practicality in Production} 

The choice between an open-source model and a closed-source model depends on performance, costs, and inference speed~\citep{howell_economic_2023}. Encoder-based models, such as RoBERTa-large and ModernBERT, significantly outperform decoder-based models like LLaMA 8B in terms of inference cost efficiency as shown in Table \ref{tab:acc}. While fine-tuning encoders involves initial computational overhead, this investment is quickly amortized in high-throughput applications. For instance, the fine-tuning cost of ModernBERT equates roughly to fewer than 5,000 inference queries with LLAMA3 8B, emphasizing the long-term efficiency benefits of encoder models.

% The choice between an open-source model and a closed-source model depends on performance, inference speed, and costs~\citep{howell_economic_2023}. Encoders, such as BERT, offer greater efficiency than autoregressive models, enabling faster, parallelized inference with lower computational overhead. This makes them ideal for real-time applications, ensuring scalability and cost-effectiveness.

\section{Future work}

% Interesting future research directions include considering the collective support provided by multiple documents and extending the proposed pre-filtering approach with a post-generation verification to ensure that the generated response is based on the documents provided in context. These complementary extensions have the potential to develop more factual and helpful RAG-LLM systems.

Future research should further explore how groundedness detection methods can handle scenarios involving multiple documents, where collectively supportive information must be integrated to accurately assess groundedness. Investigating mechanisms for aggregating passage-level judgments could improve multi-hop question answering, a scenario not specifically addressed in this study.

Additionally, addressing the detection of internal contradictions within retrieved contexts, a limitation of the current method, represents an important avenue for future research. Incorporating factual consistency checks could significantly enhance the robustness and reliability of groundedness detection systems.
\section{Conclusion}

This paper investigates the task of predicting whether a question is supported by the document provided in context before the computationally expensive answer generation. Our experiments reveal that lightweight encoders such as RoBERTa and NomicBERT perform well and can closely match the performance of state-of-the-art LLMs such as Llama-8B in some cases. These findings highlight the potential of encoders as an efficient approach to improving the groundedness of NLP systems.

\section*{Limitations}

While we observed that encoders perform competitively at groundedness detection on four datasets, the performance gap may widen as domains become more complex. Also, our method primarily targets single-document scenarios and straightforward question-answering tasks. It has not been evaluated on multi-hop question answering tasks that require synthesizing information from multiple documents simultaneously, such as HotpotQA or MuSiQue. In addition, our approach does not handle internal contradiction detection within the context, potentially allowing contexts containing contradictory yet relevant information to pass undetected. Thus, additional mechanisms for detecting factual consistency or contradictions would be essential for comprehensive groundedness evaluation.

\section*{Acknowledgements}

Sarath Chandar is supported by the Canada CIFAR AI Chairs program, the Canada Research Chair in Lifelong Machine Learning, and the NSERC Discovery Grant. This research was enabled in part by compute resources provided by Mila and Aily.

% Bibliography entries for the entire Anthology, followed by custom entries
%\bibliography{anthology,custom}
% Custom bibliography entries only
% \clearpage
% \bibliography{custom}

\begin{thebibliography}{27}
\providecommand{\natexlab}[1]{#1}

\bibitem[{Asai et~al.(2023)Asai, Wu, Wang, Sil, and Hajishirzi}]{asai_self-rag_2023}
Akari Asai, Zeqiu Wu, Yizhong Wang, Avirup Sil, and Hannaneh Hajishirzi. 2023.
\newblock \href {https://arxiv.org/abs/2310.11511} {Self-rag: Learning to retrieve, generate, and critique through self-reflection}.
\newblock \emph{Preprint}, arXiv:2310.11511.

\bibitem[{Asai et~al.(2024)Asai, Zhong, Chen, Koh, Zettlemoyer, Hajishirzi, and Yih}]{asai_reliable_2024}
Akari Asai, Zexuan Zhong, Danqi Chen, Pang~Wei Koh, Luke Zettlemoyer, Hannaneh Hajishirzi, and Wen-tau Yih. 2024.
\newblock \href {https://doi.org/10.48550/arXiv.2403.03187} {Reliable, adaptable, and attributable language models with retrieval}.
\newblock \emph{Preprint}, arxiv:2403.03187 [cs].

\bibitem[{Cuconasu et~al.(2024)Cuconasu, Trappolini, Siciliano, Filice, Campagnano, Maarek, Tonellotto, and Silvestri}]{cuconasu_power_2024}
Florin Cuconasu, Giovanni Trappolini, Federico Siciliano, Simone Filice, Cesare Campagnano, Yoelle Maarek, Nicola Tonellotto, and Fabrizio Silvestri. 2024.
\newblock \href {https://doi.org/10.1145/3626772.3657834} {The power of noise: Redefining retrieval for rag systems}.
\newblock In \emph{Proceedings of the 47th International ACM SIGIR Conference on Research and Development in Information Retrieval}, SIGIR 2024, page 719–729. ACM.

\bibitem[{Devlin et~al.(2019)Devlin, Chang, Lee, and Toutanova}]{devlin_bert_2019}
Jacob Devlin, Ming-Wei Chang, Kenton Lee, and Kristina Toutanova. 2019.
\newblock \href {https://arxiv.org/abs/1810.04805} {Bert: Pre-training of deep bidirectional transformers for language understanding}.
\newblock \emph{Preprint}, arXiv:1810.04805.

\bibitem[{Howell et~al.(2023)Howell, Christian, Fomitchov, Kehat, Marzulla, Rolston, Tredup, Zimmerman, Selfridge, and Bradley}]{howell_economic_2023}
Kristen Howell, Gwen Christian, Pavel Fomitchov, Gitit Kehat, Julianne Marzulla, Leanne Rolston, Jadin Tredup, Ilana Zimmerman, Ethan Selfridge, and Joseph Bradley. 2023.
\newblock \href {https://doi.org/10.18653/v1/2023.acl-industry.24} {The economic trade-offs of large language models: A case study}.
\newblock In \emph{Proceedings of the 61st Annual Meeting of the Association for Computational Linguistics (Volume 5: Industry Track)}, pages 248--267, Toronto, Canada. Association for Computational Linguistics.

\bibitem[{Hsieh et~al.(2024)Hsieh, Chuang, Li, Wang, Le, Kumar, Glass, Ratner, Lee, Krishna, and Pfister}]{hsieh_found_2024}
Cheng-Yu Hsieh, Yung-Sung Chuang, Chun-Liang Li, Zifeng Wang, Long~T. Le, Abhishek Kumar, James Glass, Alexander Ratner, Chen-Yu Lee, Ranjay Krishna, and Tomas Pfister. 2024.
\newblock \href {https://arxiv.org/abs/2406.16008} {Found in the middle: Calibrating positional attention bias improves long context utilization}.
\newblock \emph{Preprint}, arXiv:2406.16008.

\bibitem[{Ji et~al.(2023)Ji, Lee, Frieske, Yu, Su, Xu, Ishii, Bang, Madotto, and Fung}]{ji_survey_2023}
Ziwei Ji, Nayeon Lee, Rita Frieske, Tiezheng Yu, Dan Su, Yan Xu, Etsuko Ishii, Ye~Jin Bang, Andrea Madotto, and Pascale Fung. 2023.
\newblock \href {https://doi.org/10.1145/3571730} {Survey of hallucination in natural language generation}.
\newblock \emph{ACM Computing Surveys}, 55(12):1–38.

\bibitem[{Jiang et~al.(2025)Jiang, Sun, Liang, and Zhang}]{jiang2025retrievesummarizeplanadvancing}
Zhouyu Jiang, Mengshu Sun, Lei Liang, and Zhiqiang Zhang. 2025.
\newblock \href {https://arxiv.org/abs/2407.13101} {Retrieve, summarize, plan: Advancing multi-hop question answering with an iterative approach}.
\newblock \emph{Preprint}, arXiv:2407.13101.

\bibitem[{Lewis et~al.(2021)Lewis, Perez, Piktus, Petroni, Karpukhin, Goyal, Küttler, Lewis, tau Yih, Rocktäschel, Riedel, and Kiela}]{lewis_retrieval-augmented_2021}
Patrick Lewis, Ethan Perez, Aleksandra Piktus, Fabio Petroni, Vladimir Karpukhin, Naman Goyal, Heinrich Küttler, Mike Lewis, Wen tau Yih, Tim Rocktäschel, Sebastian Riedel, and Douwe Kiela. 2021.
\newblock \href {https://arxiv.org/abs/2005.11401} {Retrieval-augmented generation for knowledge-intensive nlp tasks}.
\newblock \emph{Preprint}, arXiv:2005.11401.

\bibitem[{Liu et~al.(2023)Liu, Lin, Hewitt, Paranjape, Bevilacqua, Petroni, and Liang}]{liu_lost_2023}
Nelson~F. Liu, Kevin Lin, John Hewitt, Ashwin Paranjape, Michele Bevilacqua, Fabio Petroni, and Percy Liang. 2023.
\newblock \href {https://arxiv.org/abs/2307.03172} {Lost in the middle: How language models use long contexts}.
\newblock \emph{Preprint}, arXiv:2307.03172.

\bibitem[{Liu et~al.(2019)Liu, Ott, Goyal, Du, Joshi, Chen, Levy, Lewis, Zettlemoyer, and Stoyanov}]{liu_roberta_2019}
Yinhan Liu, Myle Ott, Naman Goyal, Jingfei Du, Mandar Joshi, Danqi Chen, Omer Levy, Mike Lewis, Luke Zettlemoyer, and Veselin Stoyanov. 2019.
\newblock \href {https://arxiv.org/abs/1907.11692} {Roberta: A robustly optimized bert pretraining approach}.
\newblock \emph{Preprint}, arXiv:1907.11692.

\bibitem[{Nussbaum et~al.(2025)Nussbaum, Morris, Duderstadt, and Mulyar}]{nussbaum_nomic_2025}
Zach Nussbaum, John~X. Morris, Brandon Duderstadt, and Andriy Mulyar. 2025.
\newblock \href {https://doi.org/10.48550/arXiv.2402.01613} {Nomic embed: Training a reproducible long context text embedder}.
\newblock \emph{Preprint}, arXiv:2402.01613 [cs].

\bibitem[{Rajpurkar et~al.(2018)Rajpurkar, Jia, and Liang}]{rajpurkar_know_2018}
Pranav Rajpurkar, Robin Jia, and Percy Liang. 2018.
\newblock \href {https://arxiv.org/abs/1806.03822} {Know what you don't know: Unanswerable questions for squad}.
\newblock \emph{Preprint}, arXiv:1806.03822.

\bibitem[{Salinas and Morstatter(2024)}]{salinas_butterfly_2024}
Abel Salinas and Fred Morstatter. 2024.
\newblock \href {https://doi.org/10.18653/v1/2024.findings-acl.275} {The butterfly effect of altering prompts: How small changes and jailbreaks affect large language model performance}.
\newblock In \emph{Findings of the Association for Computational Linguistics: {ACL} 2024}, pages 4629--4651, Bangkok, Thailand. Association for Computational Linguistics.

\bibitem[{Thakur et~al.(2021)Thakur, Reimers, Rücklé, Srivastava, and Gurevych}]{thakur_beir_2021}
Nandan Thakur, Nils Reimers, Andreas Rücklé, Abhishek Srivastava, and Iryna Gurevych. 2021.
\newblock \href {https://arxiv.org/abs/2104.08663} {Beir: A heterogenous benchmark for zero-shot evaluation of information retrieval models}.
\newblock \emph{Preprint}, arXiv:2104.08663.

\bibitem[{Trischler et~al.(2017)Trischler, Wang, Yuan, Harris, Sordoni, Bachman, and Suleman}]{trischler_newsqa_2017}
Adam Trischler, Tong Wang, Xingdi Yuan, Justin Harris, Alessandro Sordoni, Philip Bachman, and Kaheer Suleman. 2017.
\newblock \href {https://arxiv.org/abs/1611.09830} {Newsqa: A machine comprehension dataset}.
\newblock \emph{Preprint}, arXiv:1611.09830.

\bibitem[{Wang et~al.(2024{\natexlab{a}})Wang, Ren, Li, Zhao, Liu, and Wen}]{wang_rear_2024}
Yuhao Wang, Ruiyang Ren, Junyi Li, Wayne~Xin Zhao, Jing Liu, and Ji-Rong Wen. 2024{\natexlab{a}}.
\newblock \href {https://arxiv.org/abs/2402.17497} {Rear: A relevance-aware retrieval-augmented framework for open-domain question answering}.
\newblock \emph{Preprint}, arXiv:2402.17497.

\bibitem[{Wang et~al.(2024{\natexlab{b}})Wang, Ren, Li, Zhao, Liu, and Wen}]{wang-etal-2024-rear}
Yuhao Wang, Ruiyang Ren, Junyi Li, Xin Zhao, Jing Liu, and Ji-Rong Wen. 2024{\natexlab{b}}.
\newblock \href {https://doi.org/10.18653/v1/2024.emnlp-main.321} {{REAR}: A relevance-aware retrieval-augmented framework for open-domain question answering}.
\newblock In \emph{Proceedings of the 2024 Conference on Empirical Methods in Natural Language Processing}, pages 5613--5626, Miami, Florida, USA. Association for Computational Linguistics.

\bibitem[{Wang et~al.(2023)Wang, Araki, Jiang, Parvez, and Neubig}]{wang_learning_2023}
Zhiruo Wang, Jun Araki, Zhengbao Jiang, Md~Rizwan Parvez, and Graham Neubig. 2023.
\newblock \href {https://doi.org/10.48550/arXiv.2311.08377} {Learning to filter context for retrieval-augmented generation}.
\newblock \emph{Preprint}, arxiv:2311.08377 [cs].

\bibitem[{Wu et~al.(2024)Wu, Xie, Chen, Zhu, Zhang, and Xiao}]{wu_how_2024}
Siye Wu, Jian Xie, Jiangjie Chen, Tinghui Zhu, Kai Zhang, and Yanghua Xiao. 2024.
\newblock \href {https://arxiv.org/abs/2404.03302} {How easily do irrelevant inputs skew the responses of large language models?}
\newblock \emph{Preprint}, arXiv:2404.03302.

\bibitem[{Xie et~al.(2024)Xie, Zhang, Chen, Lou, and Su}]{xie_adaptive_2024}
Jian Xie, Kai Zhang, Jiangjie Chen, Renze Lou, and Yu~Su. 2024.
\newblock \href {https://arxiv.org/abs/2305.13300} {Adaptive chameleon or stubborn sloth: Revealing the behavior of large language models in knowledge conflicts}.
\newblock \emph{Preprint}, arXiv:2305.13300.

\bibitem[{Yoran et~al.(2024)Yoran, Wolfson, Ram, and Berant}]{yoran_making_2024}
Ori Yoran, Tomer Wolfson, Ori Ram, and Jonathan Berant. 2024.
\newblock \href {https://arxiv.org/abs/2310.01558} {Making retrieval-augmented language models robust to irrelevant context}.
\newblock \emph{Preprint}, arXiv:2310.01558.

\bibitem[{Zhang et~al.(2024)Zhang, Zhang, Zhang, Liu, and Huang}]{zhang_end--end_2024}
Jiahao Zhang, Haiyang Zhang, Dongmei Zhang, Yong Liu, and Shen Huang. 2024.
\newblock \href {https://arxiv.org/abs/2308.08973} {End-to-end beam retrieval for multi-hop question answering}.
\newblock \emph{Preprint}, arXiv:2308.08973.

\bibitem[{Zhang et~al.(2023)Zhang, Deng, Liu, Pan, and Bing}]{zhang_sentiment_2023}
Wenxuan Zhang, Yue Deng, Bing Liu, Sinno~Jialin Pan, and Lidong Bing. 2023.
\newblock \href {https://arxiv.org/abs/2305.15005} {Sentiment analysis in the era of large language models: A reality check}.
\newblock \emph{Preprint}, arXiv:2305.15005.

\bibitem[{Zhao et~al.(2024)Zhao, Zhou, Li, Tang, Wang, Hou, Min, Zhang, Zhang, Dong, Du, Yang, Chen, Chen, Jiang, Ren, Li, Tang, Liu, Liu, Nie, and Wen}]{zhao_survey_2024}
Wayne~Xin Zhao, Kun Zhou, Junyi Li, Tianyi Tang, Xiaolei Wang, Yupeng Hou, Yingqian Min, Beichen Zhang, Junjie Zhang, Zican Dong, Yifan Du, Chen Yang, Yushuo Chen, Zhipeng Chen, Jinhao Jiang, Ruiyang Ren, Yifan Li, Xinyu Tang, Zikang Liu, Peiyu Liu, Jian-Yun Nie, and Ji-Rong Wen. 2024.
\newblock \href {https://doi.org/10.48550/arXiv.2303.18223} {A survey of large language models}.
\newblock \emph{Preprint}, arXiv:2303.18223 [cs].

\bibitem[{Zhou et~al.(2024)Zhou, Liu, Jin, Nie, and Dou}]{zhou2024metacognitiveretrievalaugmentedlargelanguage}
Yujia Zhou, Zheng Liu, Jiajie Jin, Jian-Yun Nie, and Zhicheng Dou. 2024.
\newblock \href {https://arxiv.org/abs/2402.11626} {Metacognitive retrieval-augmented large language models}.
\newblock \emph{Preprint}, arXiv:2402.11626.

\bibitem[{Zimmerman et~al.(2024)Zimmerman, Tredup, Selfridge, and Bradley}]{zimmerman_two-tiered_2024}
Ilana Zimmerman, Jadin Tredup, Ethan Selfridge, and Joseph Bradley. 2024.
\newblock \href {https://doi.org/10.18653/v1/2024.emnlp-industry.2} {Two-tiered encoder-based hallucination detection for retrieval-augmented generation in the wild}.
\newblock In \emph{Proceedings of the 2024 Conference on Empirical Methods in Natural Language Processing: Industry Track}, pages 8--22, Miami, Florida, US. Association for Computational Linguistics.

\end{thebibliography}
\bibliographystyle{plainnat}

\appendix
\section{Appendix}
\label{sec:appendix}

\subsection{Hyperparameter Tuning} \label{sec:hyperparameter}
To ensure the robustness and reproducibility of our results, we conducted extensive hyperparameter tuning using grid search. The tuning process explored various configurations of key parameters, including learning rates, batch sizes, and regularization techniques. Specifically, we experimented with different learning rates, batch sizes, weight decay values, and learning rate scheduling strategies. The hyperparameter search space is summarized in Table~\ref{tab:hyperparameters}.

All experiments were conducted using consistent data preprocessing pipelines and evaluation metrics to facilitate a fair comparison between model architectures. This systematic approach ensured that the selected hyperparameters yielded optimal performance across different datasets while maintaining reproducibility.

\begin{table}[htb]
	\centering
	\begin{tabular}{|c|c|}
		\hline
		\textbf{Hyperparameter} & \textbf{Values}        \\
		\hline
		Learning Rate           & \{1.5e-5, 2e-5, 3e-5\} \\
		\hline
		Batch Size              & \{8, 16, 32\}          \\
		\hline
		Weight Decay            & \{0.1, 0.01\}          \\
		\hline
		Learning Rate Scheduler & \{Linear, Cosine\}     \\
		\hline
		Warm-up Ratio           & \{0.06, 0.25\}         \\
		\hline
	\end{tabular}
	\caption{Hyperparameter search space used in the experiments.}
	\label{tab:hyperparameters}
\end{table}

% By adopting this comprehensive evaluation framework, we aim to gain deeper insights into the strengths and limitations of encoder-based and decoder-based models in handling groundedness classification tasks across diverse contexts.

\subsection{Prompts used for our setting} \label{sec:prompts}
We've experimented with the following prompts, for the NewsQA and SquadV2 datasets we used: 
\begin{itemize}
	\item Can you answer the question using the given context? Reply with 'yes' or 'no'.
	\item Based solely on the provided context, is the question answerable? Respond 'yes' or 'no'.
	\item Evaluate the question with the given context. Can the context provide an answer? Reply 'yes' or 'no'.
	\item Verify if the question can be answered using the context. Answer with 'yes' or 'no'.
	\item Is the question answerable from the context provided? Answer 'yes' or 'no'.
	\item Determine if the context provides enough information to answer the question. Respond 'yes' or 'no'.
	\item Assess the context and determine if it answers the question. Reply with 'yes' or 'no'.
	\item Given the context, decide if the question can be answered. Respond 'yes' or 'no'.
	\item Does the context contain sufficient information to answer the question? Reply with 'yes' or 'no'.
	\item Based on the context, is it possible to answer the question? Answer 'yes' or 'no'.
	\item Examine the context and decide if it answers the question. Respond with 'yes' or 'no'.
	\item Evaluate the given context to determine if the question can be answered. Reply 'yes' or 'no'.
	\item Analyze the context and determine if it provides an answer to the question. Respond 'yes' or 'no'.
	\item Does the context provide an answer to the question? Answer 'yes' or 'no'.
	\item Evaluate if the context answers the question. Reply with 'yes' or 'no'.
	\item Is there enough information in the context to answer the question? Respond 'yes' or 'no'.
	\item Analyze the context and decide if it sufficiently answers the question. Reply 'yes' or 'no'.
	\item Based on the context, determine if the question is answerable. Answer 'yes' or 'no'.
	\item Verify whether the context answers the question. Reply with 'yes' or 'no'.
	\item Using only the context provided, decide if you can answer the question. Respond 'yes' or 'no'.
	          
\end{itemize}
For information retrieval datasets we used :

\begin{itemize}
	\item Does the context provide relevant information to answer the query? Respond with 'yes' or 'no'.
	\item Based on the context, is the information provided relevant to answering the query? Answer 'yes' or 'no'.
	\item Assess whether the context contains relevant details to answer the query. Reply with 'yes' or 'no'.
	\item Evaluate if the context is relevant to the query. Respond with 'yes' or 'no'.
	\item Can the context help in answering the query? Respond with 'yes' or 'no'.
	\item Analyze the relevance of the context and the query. Answer 'yes' or 'no'.
	\item Determine if the context contains pertinent information to answer the question. Reply with 'yes' or 'no'.
	\item Does the context include relevant information to address the question? Respond 'yes' or 'no'.
	\item Evaluate whether the context is closely related to the query. Reply with 'yes' or 'no'.
	\item Based on the context, assess if the details are relevant for answering the question. Respond with 'yes' or 'no'.
	\item Determine if the context is sufficient to answer the query. Respond with 'yes' or 'no'.
	\item Assess whether the context directly addresses the query. Answer 'yes' or 'no'.
	\item Does the context contain enough information to respond to the query? Reply with 'yes' or 'no'.
	\item Analyze the context and decide if it is relevant to the query. Respond with 'yes' or 'no'.
	\item Check if the context provides a direct answer to the query. Reply with 'yes' or 'no'.
	\item Evaluate the extent to which the context relates to the query. Respond with 'yes' or 'no'.
	\item Determine whether the query can be answered based on the given context. Answer 'yes' or 'no'.
	\item Is the context aligned with the information needed to answer the query? Respond 'yes' or 'no'.
	\item Judge if the context contains meaningful details to answer the question. Reply with 'yes' or 'no'.
	\item Decide whether the context provides necessary information to answer the query. Respond with 'yes' or 'no'.
\end{itemize}

\end{document}